\pdfoutput=1

\documentclass[11pt]{article}

\usepackage{EMNLP2023}

\usepackage{times}
\usepackage{latexsym}

\usepackage[T1]{fontenc}

\usepackage[utf8]{inputenc}

\usepackage{microtype}

\usepackage{inconsolata}

\usepackage{url}
\usepackage{graphicx}

\usepackage{fancyvrb}
\usepackage{color}
\newcommand\verbbf[1]{\textcolor[rgb]{0,0,1}{\textbf{#1}}}

\usepackage{todonotes}

\title{Dialogue-based generation of self-driving simulation scenarios using Large Language Models}

\author{Antonio Valerio Miceli-Barone$^1$ \\
  \texttt{amiceli@ed.ac.uk} \\\And
  Alex Lascarides$^1$ \\
  \texttt{alex@inf.ed.ac.uk} \\\And
  Craig Innes$^1$ \\
  \texttt{craig.innes@ed.ac.uk}
  \AND
\textnormal{\normalsize $^1$ School of Informatics, University of Edinburgh}
  }

\begin{document}
\maketitle
\begin{abstract}
Simulation is an invaluable tool for developing and evaluating controllers for self-driving cars. Current simulation frameworks are driven by highly-specialist domain specific languages, and so a natural language interface would greatly enhance usability.  But there is often a gap, consisting of tacit assumptions the user is making, between a concise English utterance and the executable code that captures the user's intent.  In this paper we describe a system that addresses this issue by supporting an extended multimodal interaction: the user can follow up prior instructions with refinements or revisions, in reaction to the simulations that have been generated from their utterances so far.  We use Large Language Models (LLMs) to map the user's English utterances in this interaction into domain-specific code, and so we explore the extent to which LLMs capture the context sensitivity that's necessary for computing the speaker's intended message in discourse.
\end{abstract}

\section{Introduction}
\label{SEC:INTRODUCTION}

Developing self-driving vehicles is a highly complicated process.  In particular, a  long tail of edge cases, in which the controller's current policies are likely to yield bad outcomes, needs to be discovered and tackled.

Testing self-driving vehicles on public roads is expensive and potentially dangerous \citep{bbc2018ubercrash}: safety-critical near-crash scenarios occur with low probability during normal operation, and for ethical reasons they cannot be created deliberately.
Testing on closed circuits or other artificial environments is only a partial solution since it cannot easily account for the high variability of conditions on public roads.
Simulation is a potentially cheaper, faster and safer alternative that can complement road testing.
It allows developers to evaluate rare, dangerous or even counterfactual scenarios (e.g. scenarios that involve road configurations or vehicle hardware components that do not yet exist).
Furthermore, some simulation methods may be amenable to formal verification.

However, realistic simulation is hard.
Over the years, many simulation frameworks for self-driving have been developed, each focusing on different aspects and hence different abstraction levels of the task \citep{Hoss2021Review}.
Each simulation framework is accompanied with an application programming interface (API) or a domain specific language (DSL), so that users can specify scenarios and control algorithms programmatically.
Despite the ongoing standardization efforts \citep{ASAM2020OpenODD, ASAM2020OpenXOntology, Thorn2018AFF, PAS_1883_2020}, different frameworks still use different interfaces, ontologies and paradigms.
This creates a steep learning curve for engineers. 
Ideally, we would like to have systems that allow engineers to specify simulation scenarios with minimal specific knowledge of these details, using natural language as an interface.

Recently, coding assistants based on Large Language Models (LLMs) have become good enough to be of practical use \citep{GitHubCopilot2021, Nijkamp2022codegen, GitHubCopilotX2023, openai2023gpt4}.
In this approach, the user specifies programming problems by providing natural language instructions and optionally partial programs, and the assistant provides program completions.
The most recent systems, such as GitHub Copilot X, OpenAI ChatGPT/GPT-4 and Anthropic Claude, allow multiple rounds of interaction with the user through dialogue, enabling an iterative refinement of solutions.
These systems usually perform well on common programming languages such as Python (though they can fail systematically on unusual coding tasks that require non-trivial reasoning \citep{micelibarone2023larger}).
It would be desirable to leverage them to aid the generation of scenario specifications for self-driving vehicle evaluation from natural language descriptions.
This is a non-trivial task, since in contrast to Python, it is unlikely that these large pretrained models have seen many scenario specifications in all the specific DSLs accompanying the various simulation frameworks within their training data.  Therefore, they cannot be used directly in zero-shot mode.

An additional challenge occurs because typical natural language scenario descriptions are often highly concise, with many aspects of the user's intended configuration remaining linguistically implicit. 
For example "\textit{The ego vehicle, which is in the left lane, overtakes the car in front}" does \textbf{not} make explicit that this action is intended to be not at a crossroads. 
But a simulator may randomly sample aspects of the scenario that are absent from the description, and therefore may generate a crossroads, conflicting with the user's intent.

In this project, we explore a methodology to develop a dialogue-based coding assistant for the generation of self-driving vehicle simulation scenarios using LLMs.
Our main objective is to generate Scenic scenarios \citep{Fremont2022Scenic} from English descriptions.
We focus on the NHTSA-inspired pre-crash scenarios for the Carla Autonomous Driving challenge\footnote{\url{https://carlachallenge.org/challenge/nhtsa/}}.
Due to the extremely low-resource nature of this problem (we were able to collect only $32$ examples of scenarios with an English description) we make extensive use of in-context learning techniques.
We evaluate to what extent dialogue-based interaction can contribute to this task, under the hypothesis that it allows the user to iteratively evaluate and refine the generated scenarios by identifying which implicit assumptions are being violated and making them explicit to the system. 


\section{Experiments}

\subsection{Frameworks}

\begin{figure}[h]
    \centering
    \includegraphics[width=0.49\textwidth]{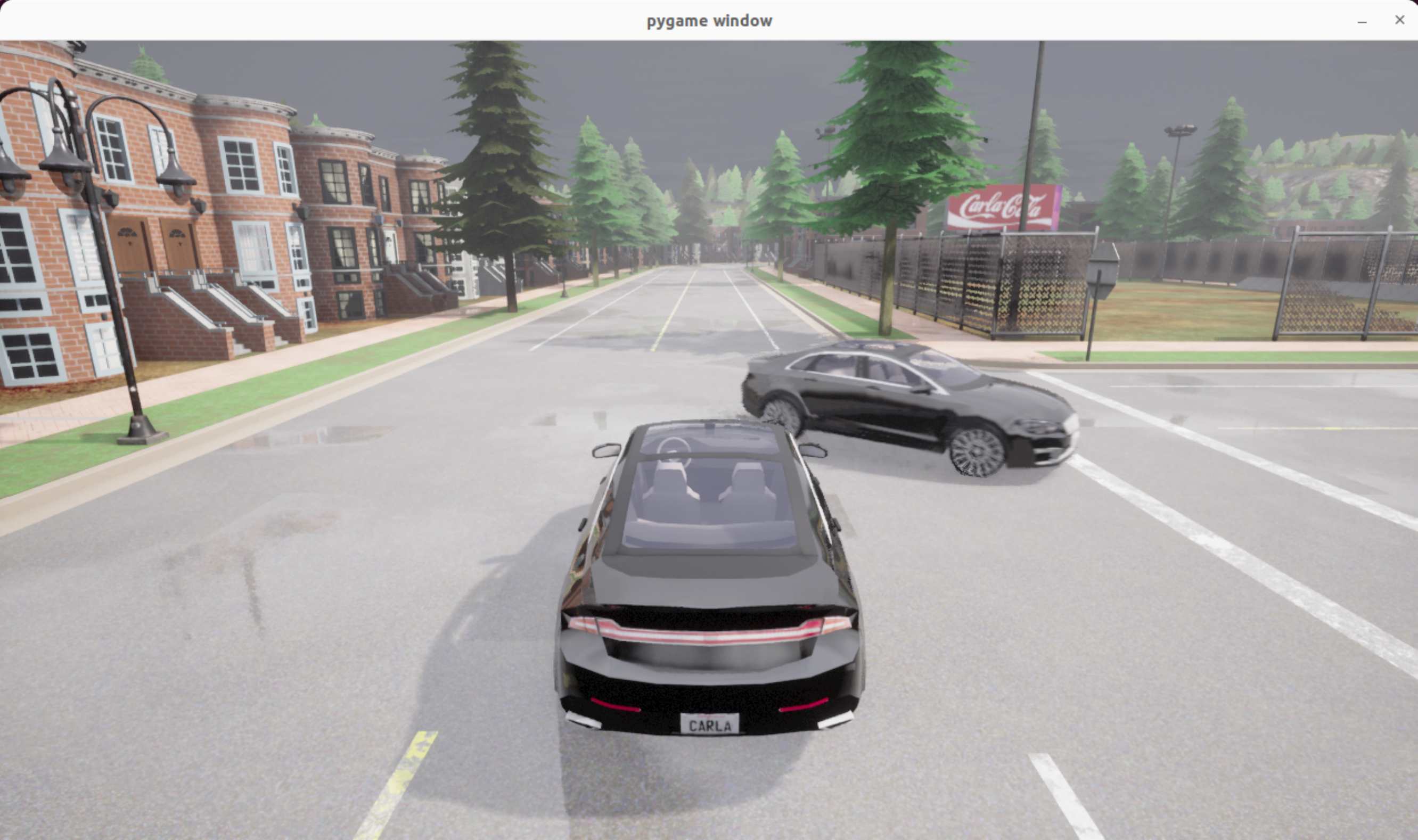}
    \caption{A Scenic scenario generated by our model and executed in the CARLA simulator. The description is: "\textit{Ego vehicle goes straight at 3-way intersection and must suddenly stop to avoid collision when adversary vehicle makes a left turn.}"}
    \label{fig:results:scenario_screenshoot}
\end{figure}

A Scenic \textit{scenario} is a probabilistic program which defines a distribution over \textit{scenes}, which specify the initial positions and velocities of vehicles, pedestrians and other objects on a map representing an urban environment, and also specify their behaviours.
This probabilistic formulation enables the concise specification of a large number of test cases, in which the implicit aspects are randomly sampled by the scenario compiler.
The sampled scenes are executed by a backend--- in our work, the CARLA Urban Driving Simulator \citep{Dosovitskiy17} (Figure~\ref{fig:results:scenario_screenshoot}).

For scenario generation from English descriptions we use the OpenAI GPT-4 model version 0314, available through the OpenAI API.\footnote{\url{https://platform.openai.com/docs/api-reference}}
GPT-4 is an instruction-tuned LLM optimized for dialogue-based interaction.

\subsection{Available data resources}

The Scenic repository has only 32 scenario specifications with both code and natural language descriptions.\footnote{\url{https://github.com/BerkeleyLearnVerify/Scenic/tree/main/examples/carla}}
We use 16 scenarios as training examples and 16 as test examples.
Each scenario description consists of one or two English sentences, mainly describing behaviours and a location type where the simulated incident happens (e.g. "on a highway", "at a 4-way crossroad", etc.), although the location may be left implicit.  
The scenario code, while being probabilistic, is more specific than the English description, as it includes ranges for initial vehicle distances and velocities, algorithmic instructions for how to locate a suitable starting point on the map, and agent behaviours specified by composing elementary behaviours in a subsumption-style architecture \citep{Brooks1986Subsumption}.
Therefore, there are two levels of underspecification here: one from the English description to the probabilistic scenario code, and another from the probabilistic code to the executable scene.
This reflects how humans communicate with language: they often leave implicit aspects of their intended message, especially when it can be inferred from what they have made explicit via commonsense inference.

\subsection{Workflow}

In our experiments, we focus on a use case where a simulation scenario is generated through multiple rounds of interaction between an engineer (the user) and the system (the assistant).

Specifically, the user first provides a short English description of the scenario consisting of one or two sentences in the style of the Carla Autonomous Driving scenarios, for instance "\textit{Ego-vehicle is driving in the left lane of a highway and must perform a lane change to avoid a slow moving or parked vehicle on the shoulder.}".
The assistant queries the instruction-tuned LLM to generate the initial code for a Scenic scenario that corresponds to the meaning of the description. 
We then invoke the Scenic compiler, which will attempt to parse the code, randomly sample a particular scene and run it in the CARLA simulator.
Since the LLM decoding process is unconstrained, the generated code may be syntactically incorrect or it may reference unknown behaviours, functions, or asset names, making the code non-executable.\footnote{Scenario execution can also fail if the rejection sampling algorithm used by Scenic fails to generate a scene that satisfies the specified constraints within a predefined number of iterations, namely 2,000.}
These issues will be detected when execution is attempted, in which case Scenic returns an exception.
In such cases, the exception message is sent back to the LLM, concatenated to the previous generation and all its prompts (up to the length limit of the LLM context window).
This process is iterated, without user intervention, until an executable scenario is produced or a maximum number of rounds (specifically, 5) is reached.
If this initial phase is successful, the user observes both the code and simulation runs of multiple scenes.
The user can then judge whether they adhere to the requirements, and if necessary ask the assistant for further variations.  The user can then express further content, and the process iterates.

For instance, assume that in the aforementioned example scenario, the assistant generated code instantiates two slow-moving vehicles rather than one, which violates Gricean scalar implicatures of the description \cite{grice:1975} (i.e., it implicates there is only one slow moving or parked vehicle) and interferes with the termination condition\footnote{This example is taken from an exploratory experiment that we executed.
}.
The user may comment "\textit{Create only one vehicle besides the ego vehicle.}".
This comment is appended to the LLM context and a new scenario is generated, performing automatic iteration if exceptions are detected, until another executable scenario is presented to the user.

At this time, for instance, the user may notice that a crash occurs in some of the runs because the ego-vehicle is driving too fast, and they may comment "\textit{Execute the lane change at a speed no more than 10.}", and the assistant may respond by changing the parameter that controls the ego-vehicle's speed in the scenario code.
If at some point during the process the assistant fails to generate an executable scenario within the specified maximum of rounds, the user can examine the generated code and provide hints to fix it back to the LLM.
Hopefully, the user will eventually be satisfied with the generated scenario.


\subsection{LLM interaction}

In order to generate scenario programs from their English descriptions, we use in-context few-shot learning \citep{brown2020language}.
For the first query of a given test scenario, we randomly sample training examples up the a total of 6,500 out of 8,000 tokens of the GPT-4 context window.
Since training examples are on average 512 tokens long, this results in about 12 examples out of 16 being selected.
After appending the test scenario English description, we query the model with softmax temperature 0.1.
We extract the code from the model response and post-process it before sending it to the Scenic compiler.
See appendix \ref{sec:appendix:prompt_templates} for details on the prompting templates and code processing.

For each the following queries, the model output message and then the Scenic exception or the user comment are appended to the prompt and training examples are removed in order to keep the prompt length under 6,500 tokens.

\section{Evaluation}

\begin{figure*}[h]
    \centering
    \includegraphics[width=0.49\linewidth]{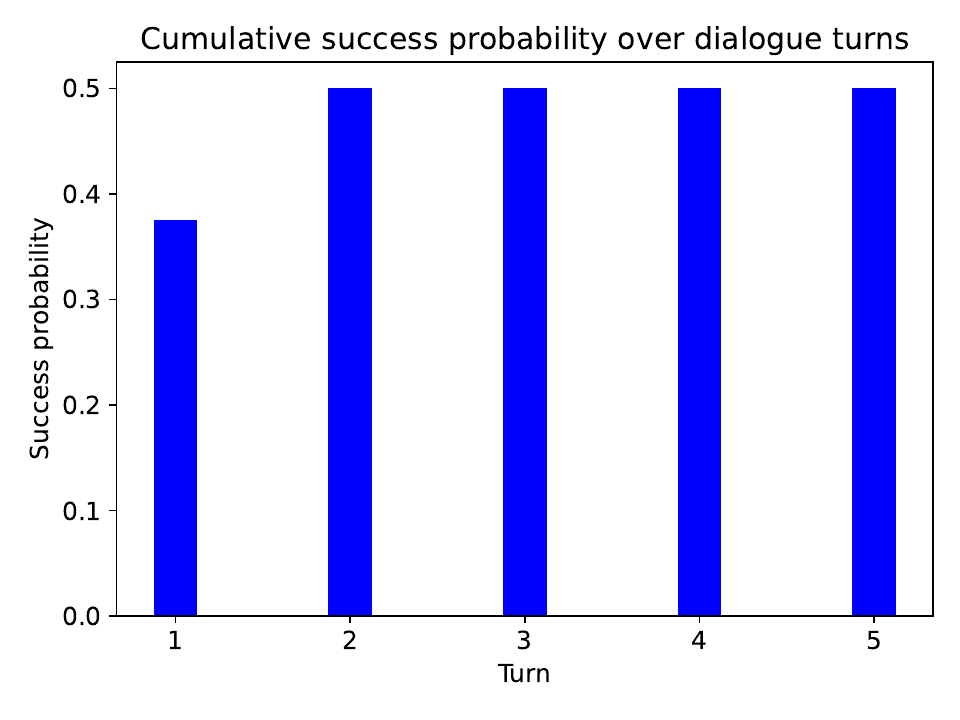}
    \includegraphics[width=0.49\linewidth]{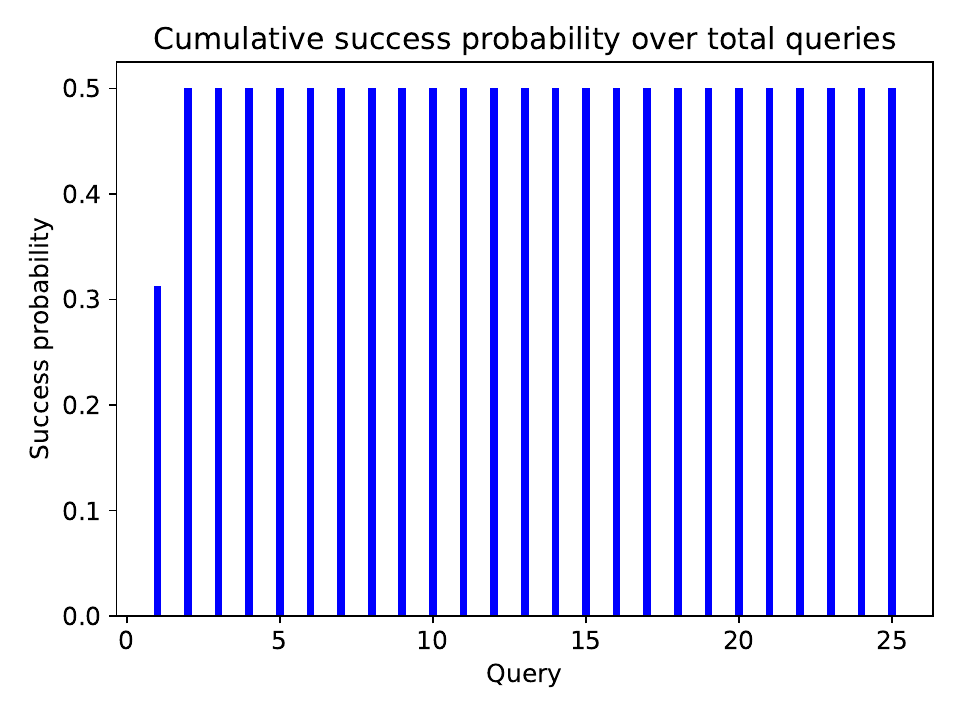}
    \caption{Cumulative success probability over number of dialogue turns (left), and over total number of LLM queries (right).}
    \label{fig:results:success_cdf}
\end{figure*}

In our evaluation protocol, we allow for a maximum of 5 dialogue turns per test scenario, including the initial generation.
For each turn, we allow up to 5 LLM queries.
When an executable scenario is generated, the user views multiple scenes sampled by Scenic until they are satisfied (scenes usually last 10-30 seconds), then the user either declares the scenario a success, or provides a comment to the system until the maximum number of turns is reached.
The user can also view the generated code at each turn, but not the gold-standard reference code.
See appendix \ref{SEC:APPENDIX:GENERATION_EXAMPLE} for an example of scenario generation with user interaction.

In this work, one of the authors acts as the user.
The OpenAI API cost for running our experiments was 19.12 USD.
The code to replicate our experiments is available on GitHub\footnote{\url{https://github.com/Avmb/DialogLLMScenic.git}}.

\subsection{Results}

We generate 8 successful scenarios out of 16 (50\%).
We report the cumulative success probability over the number of dialogue turns and over the total number of LLM queries in Figure~\ref{fig:results:success_cdf}.

We observe that two turns of dialogue increase success probability from 37.5\% of single turn generation to 50\%.
When the system successfully generates a scenario, it does so within two LLM queries, whether they happen in the first or the second turn, with success probability increasing from 31.3\% to 50\%.
Additional turns or queries do not improve success probability, which is somewhat disappointing, although even with one turn we obtain a substantial gain over the direct generation baseline.

Qualitative analysis reveals that the system is often able to generate executable code within one or two queries per turn, but unless it generates the correct scenario semantics early on, it tends to struggle with details such as the relative movement direction of cars and pedestrians.
In a small number of cases, the system keeps generating incorrect keyword or identifier names (essentially trying to guess the syntax and standard library of Scenic) or generates scenarios that can compile but fail in the sampling stage due to excessively tight constraints that cause the Scenic rejection sampler to run out of iterations.
In these cases, the user can sometimes correct the model by looking at the generated code and providing suitable hints.
See appendix \ref{SEC:APPENDIX:GENERATION_EXAMPLE} for examples of both a successful and a failed scenario generation.

\section{Conclusions}

We investigated the generation of stochastic self-driving simulation scenarios using dialogue-based interaction between an user and a LLM. 
We considered a very challenging task due to the very small number of training examples and the use of a domain-specific language which the LLM has probably not seen neither during its pre-training nor its instruction fine-tuning.

Using iterative model querying and multiple turns of interaction with the user we were able to achieve success in half of our test scenarios, a substantial improvement over the direct single-query generation baseline.
While the user still need expertise in the scenario specification language, our approach may allow faster prototyping compared to writing scenarios from scratch.

We believe that this work constitutes a promising research direction in the field of natural language interfaces for self-driving vehicle engineering and, more generally, robotics.
Future work may extend this approach to semi-automatic scenario validation or formal verification, where algorithmic evaluation criteria are also generated by the LLM through user interaction.


\nocite{Ando2005,borschinger-johnson-2011-particle,andrew2007scalable,rasooli-tetrault-2015,goodman-etal-2016-noise,harper-2014-learning}

\section*{Limitations}
\begin{itemize}
    \item Due to time constraints, our human evaluation was conducted with only one test subject which is also an author of this work. This might introduce bias. A better approach would involve multiple independent evaluators.
    \item Our approach uses a simple few-shot in-context learning setup with random example selection. More advanced LLM example selection techniques (e.g. similarity-based), possibly combined with example augmentation or more advanced prompting strategies (e.g. chain-of-thought) might lead to better performance.
    \item The LLM processing of error messages and user comments is zero-shot. Creating a dataset of such interactions obtained on the training examples and using them as additional context during deployment might increase performance.
    \item We only evaluated a single, closed-source, API-only, model.
    Better performance and better usability might be achieved on open source models which may be fine-tuned on our training examples.
\end{itemize}

\section*{Ethics Statement}

Our experiments involve the participation of a human subject (one of the authors).
Due to the nature of the experiments, we believe that there were no plausible risks for the human subject.

Our work may enable a more extensive use of simulation in the development of self-driving vehicles.
This can have positive implications for public safety, as it may reduce the amount of testing on public roads of early-stage prototypes which may endanger the public (as in the case of the Uber self-driving car crash).
However, it should be remarked that simulation can never fully replace real-world testing and post-marketing monitoring.

Our work uses a closed-source LLM which is available only as an API which may be discontinued at any time and which has been trained on unknown data.
There may be intrinsic ethical issues with such models due to their extreme black-box nature.
We recommend that production systems should be based on open-source models that can be independently audited and can be run on local hardware.

\section*{Acknowledgements}
This work was supported by a grant from the UKRI Strategic Priorities
Fund to the UKRI Research Node on Trustworthy Autonomous Systems
Governance and Regulation (EP/V026607/1, 2020-2024).

\bibliography{anthology,custom}
\bibliographystyle{acl_natbib}

\appendix

\section{Prompt templates and data processing}
\label{sec:appendix:prompt_templates}

\subsection{Templates}
For each scenario, in the initial query to the LLM, we start with a message in the GPT-4 "\textbf{system}" conversation role, providing an instruction describing the task:
"\textit{You are a helpful agent that generates specifications for car driving scenarios in the Scenic language.\\
Scenic is a domain-specific probabilistic programming language for modeling the environments of cyber-physical systems like robots and autonomous cars. A Scenic program defines a distribution over scenes, configurations of physical objects and agents; sampling from this distribution yields concrete scenes which can be simulated to produce training or testing data. Scenic can also define (probabilistic) policies for dynamic agents, allowing modeling scenarios where agents take actions over time in response to the state of the world.
\\
Your task is to generate Scenic scenarios, each according to its corresponding description in English included as a docstring. Write each scenario in a separate code box. Follow the examples below:}"\footnote{This text has includes the first paragraph of the homepage of the Scenic documentation website: \url{https://scenic-lang.readthedocs.io/en/latest/}}

Then we provide a sequence of training examples, each specified by a message in the "\textbf{example\_user}" role, which contains the English description of the scenario extracted from its docstring, followed by a message in the "\textbf{example\_assistant}" role which contains the scenario code (without the docstring) inside a code block with "scenic" language tag.
Finally we provide the English description of the test scenario in a message in the "\textbf{user}" role.
The model responds to this query with a message in the "\textbf{assistant}" role.
We search for a code block with "scenic" language tag in this message and we extract the code inside it.

For subsequent queries, we append to the prompt the "\textbf{assistant}" message generated by the model, followed by a message in the "\textbf{user}" role containing either the Scenic exception prefixed with "\textit{An error has occurred:}", or the user comment prefixed with "\textit{Comment:}".
We drop examples from the top (while keeping the "\textbf{system}" message) in order to keep the prompt length to less than 6,500 tokens.
The prompt is retained and updated between queries, effectively representing the dialogue state.

\subsection{Data processing}
We perform minimal data regex-based pre-processing on the scenario code before sending it to the LLM: we remove comments, empty lines, we fix the vehicle asset name and simulator backend names (which are sometimes incorrect due to version mismatch between Scenic and CARLA) and we remove the map filename (it is a waste of tokens which can be reconstructed from the map name).
Similarly, on the generated code we add the original English description as a docstring and we add the map filename derived from the map name. 

\newpage
\onecolumn
\section{Generation example}
\label{SEC:APPENDIX:GENERATION_EXAMPLE}

\subsection{Successful generation}
\label{SEC:APPENDIX:GENERATION_EXAMPLE:SUCCESS}
In this example, the user starts with the scenario description: "\textit{Ego vehicle makes a right turn at 4-way intersection while adversary vehicle from lateral lane goes straight.}"\\
The model generates the following code, which is immediately executable:

\begin{small}
\begin{verbatim}
param map = localPath('Scenic/tests/formats/opendrive/maps/CARLA/Town05.xodr')
param carla_map = 'Town05'
model scenic.simulators.carla.model
MODEL = 'vehicle.lincoln.mkz_2017'
EGO_INIT_DIST = [20, 25]
param EGO_SPEED = VerifaiRange(7, 10)
param EGO_BRAKE = VerifaiRange(0.5, 1.0)
ADV_INIT_DIST = [10, 15]
param ADV_SPEED = VerifaiRange(7, 10)
param SAFETY_DIST = VerifaiRange(10, 20)
CRASH_DIST = 5
TERM_DIST = 70
behaviour EgoBehavior(trajectory):
        try:
                do FollowTrajectoryBehavior(target_speed=globalParameters.EGO_SPEED, trajectory=trajectory)
        interrupt when withinDistanceToAnyObjs(self, globalParameters.SAFETY_DIST):
                take SetBrakeAction(globalParameters.EGO_BRAKE)
        interrupt when withinDistanceToAnyObjs(self, CRASH_DIST):
                terminate
intersection = Uniform(*filter(lambda i: i.is4Way, network.intersections))
egoInitLane = Uniform(*intersection.incomingLanes)
egoManeuver = Uniform(*filter(lambda m: m.type is ManeuverType.RIGHT_TURN, egoInitLane.maneuvers))
egoTrajectory = [egoInitLane, egoManeuver.connectingLane, egoManeuver.endLane]
egoSpawnPt = OrientedPoint in egoInitLane.centerline
advManeuver = Uniform(*filter(lambda m: m.type is ManeuverType.STRAIGHT, egoManeuver.conflictingManeuvers))
advInitLane = advManeuver.startLane
advTrajectory = [advInitLane, advManeuver.connectingLane, advManeuver.endLane]
advSpawnPt = OrientedPoint in advInitLane.centerline
ego = Car at egoSpawnPt,
        with blueprint MODEL,
        with behaviour EgoBehavior(egoTrajectory)
adversary = Car at advSpawnPt,
        with blueprint MODEL,
        with behaviour FollowTrajectoryBehavior(target_speed=globalParameters.ADV_SPEED, trajectory=advTrajectory)
require EGO_INIT_DIST[0] <= (distance to intersection) <= EGO_INIT_DIST[1]
require ADV_INIT_DIST[0] <= (distance from adversary to intersection) <= ADV_INIT_DIST[1]
terminate when (distance to egoSpawnPt) > TERM_DIST
\end{verbatim}
\end{small}

While this code largely captures the intended behaviour, the user notices that sometimes it results in near crashes between the two vehicles.
Therefore the user starts a second turn of dialogue with the comment: "\textit{Use a higher safety distance}".
The system responds with the code:
\newpage

\begin{small}
\begin{Verbatim}[commandchars=\\\{\}]
param map = localPath('Scenic/tests/formats/opendrive/maps/CARLA/Town05.xodr')
param carla_map = 'Town05'
model scenic.simulators.carla.model
MODEL = 'vehicle.lincoln.mkz_2017'
EGO_INIT_DIST = [20, 25]
param EGO_SPEED = VerifaiRange(7, 10)
param EGO_BRAKE = VerifaiRange(0.5, 1.0)
ADV_INIT_DIST = [10, 15]
param ADV_SPEED = VerifaiRange(7, 10)
\verbbf{param SAFETY_DIST = VerifaiRange(15, 25)  # Increased safety distance}
CRASH_DIST = 5
TERM_DIST = 70
behaviour EgoBehavior(trajectory):
        try:
                do FollowTrajectoryBehavior(target_speed=globalParameters.EGO_SPEED, trajectory=trajectory)
        interrupt when withinDistanceToAnyObjs(self, globalParameters.SAFETY_DIST):
                take SetBrakeAction(globalParameters.EGO_BRAKE)
        interrupt when withinDistanceToAnyObjs(self, CRASH_DIST):
                terminate
intersection = Uniform(*filter(lambda i: i.is4Way, network.intersections))
egoInitLane = Uniform(*intersection.incomingLanes)
egoManeuver = Uniform(*filter(lambda m: m.type is ManeuverType.RIGHT_TURN, egoInitLane.maneuvers))
egoTrajectory = [egoInitLane, egoManeuver.connectingLane, egoManeuver.endLane]
egoSpawnPt = OrientedPoint in egoInitLane.centerline
advManeuver = Uniform(*filter(lambda m: m.type is ManeuverType.STRAIGHT, egoManeuver.conflictingManeuvers))
advInitLane = advManeuver.startLane
advTrajectory = [advInitLane, advManeuver.connectingLane, advManeuver.endLane]
advSpawnPt = OrientedPoint in advInitLane.centerline
ego = Car at egoSpawnPt,
        with blueprint MODEL,
        with behaviour EgoBehavior(egoTrajectory)
adversary = Car at advSpawnPt,
        with blueprint MODEL,
        with behaviour FollowTrajectoryBehavior(target_speed=globalParameters.ADV_SPEED, trajectory=advTrajectory)
require EGO_INIT_DIST[0] <= (distance to intersection) <= EGO_INIT_DIST[1]
require ADV_INIT_DIST[0] <= (distance from adversary to intersection) <= ADV_INIT_DIST[1]
terminate when (distance to egoSpawnPt) > TERM_DIST
\end{Verbatim}
\end{small}

\begin{figure}[h]
    \centering
    \includegraphics[width=0.8\textwidth]{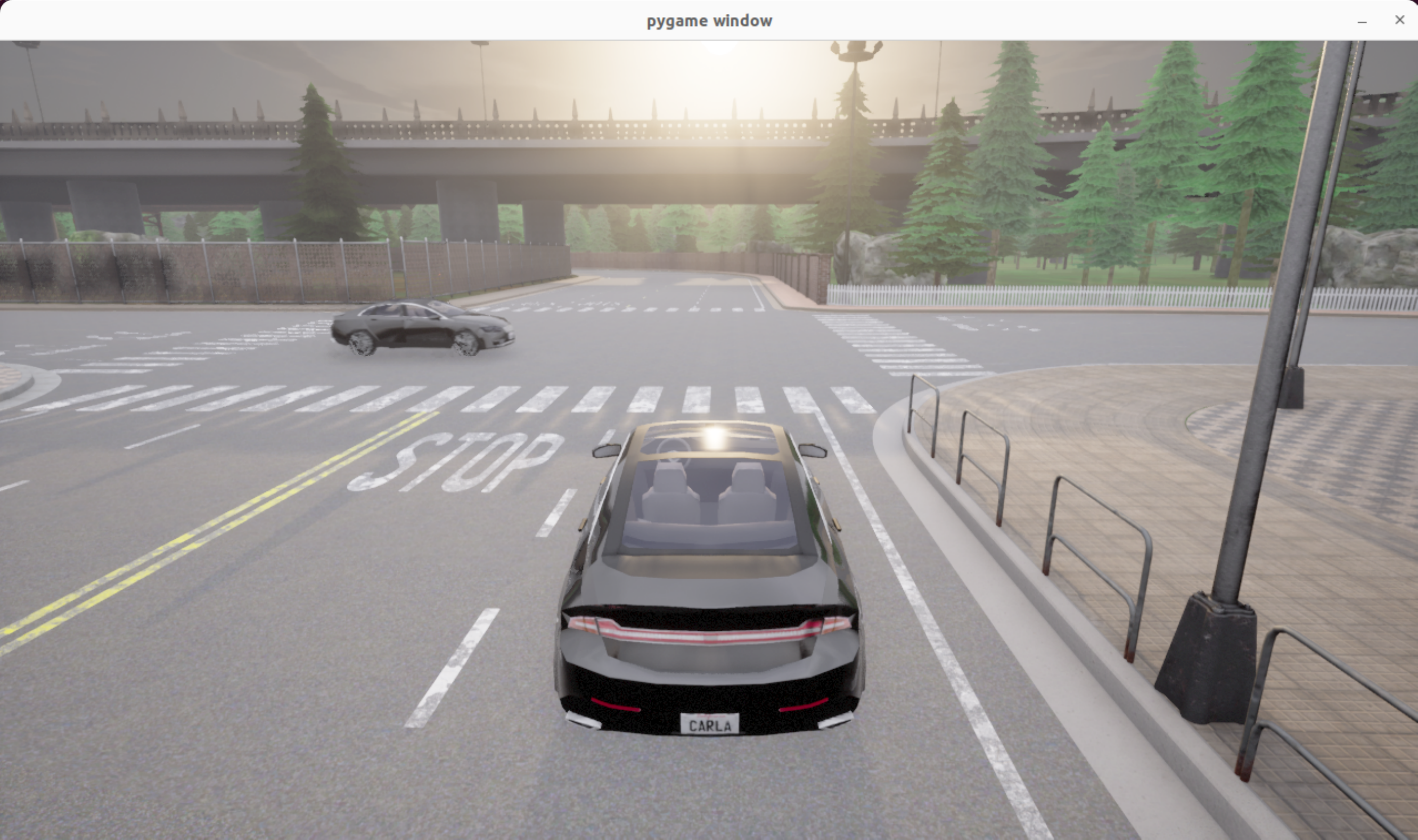}
    \caption{A scene from the successfully scenario generated in appendix \ref{SEC:APPENDIX:GENERATION_EXAMPLE:SUCCESS}.}
    \label{fig:results:scenario_screenshoot_2}
\end{figure}

The only difference with the previous generated code is the highlighted line, where the model increased the probabilistic range of possible distances to initiate the brake maneuver at.
Note that the LLM added a comment to the code to document the change, even without being instructed to.
The user observes multiple runs in CARLA simulator of scenes sampled from this code (Figure~\ref{fig:results:scenario_screenshoot_2}) and deems it satisfactory, declaring the test scenario as a success.

\subsection{Failed generation}
\label{SEC:APPENDIX:GENERATION_EXAMPLE:FAILED}
In this example, the user starts with the scenario description: "\textit{Ego vehicle makes a left turn at an intersection and must suddenly stop to avoid collision when pedestrian crosses the crosswalk.}"\\
The model generates the following code:

\begin{small}
\begin{Verbatim}[commandchars=\\\{\}]
param map = localPath('Scenic/tests/formats/opendrive/maps/CARLA/Town05.xodr')
param carla_map = 'Town05'
model scenic.simulators.carla.model
MODEL = 'vehicle.lincoln.mkz_2017'
EGO_INIT_DIST = [20, 25]
param EGO_SPEED = VerifaiRange(7, 10)
EGO_BRAKE = 1.0
PED_MIN_SPEED = 1.0
PED_THRESHOLD = 20
param SAFETY_DIST = VerifaiRange(10, 15)
CRASH_DIST = 5
TERM_DIST = 50
behaviour EgoBehavior(trajectory):
    flag = True
    try:
        do FollowTrajectoryBehavior(target_speed=globalParameters.EGO_SPEED, trajectory=trajectory)
    interrupt when withinDistanceToAnyObjs(self, globalParameters.SAFETY_DIST) and (ped in network.drivableRegion) and flag:
        flag = False
        while withinDistanceToAnyObjs(self, globalParameters.SAFETY_DIST + 3):
            take SetBrakeAction(EGO_BRAKE)
    interrupt when withinDistanceToAnyObjs(self, CRASH_DIST):
        terminate
intersection = Uniform(*filter(lambda i: i.is4Way or i.is3Way, network.intersections))
egoManeuver = Uniform(*filter(lambda m: m.type is ManeuverType.LEFT_TURN, intersection.maneuvers))
egoInitLane = egoManeuver.startLane
egoTrajectory = [egoInitLane, egoManeuver.connectingLane, egoManeuver.endLane]
egoSpawnPt = OrientedPoint in egoInitLane.centerline
tempSpawnPt = egoInitLane.centerline[-1]
ego = Car at egoSpawnPt,
    with blueprint MODEL,
    with behaviour EgoBehavior(egoTrajectory)
ped = Pedestrian right of tempSpawnPt by 5,
    with heading ego.heading,
    with regionContainedIn None,
    with behaviour CrossingBehavior(ego, PED_MIN_SPEED, PED_THRESHOLD)
require EGO_INIT_DIST[0] <= (distance to intersection) <= EGO_INIT_DIST[1]
terminate when (distance to egoSpawnPt) > TERM_DIST
\end{Verbatim}
\end{small}

\begin{figure}[h!]
    \centering
    \includegraphics[width=0.8\textwidth]{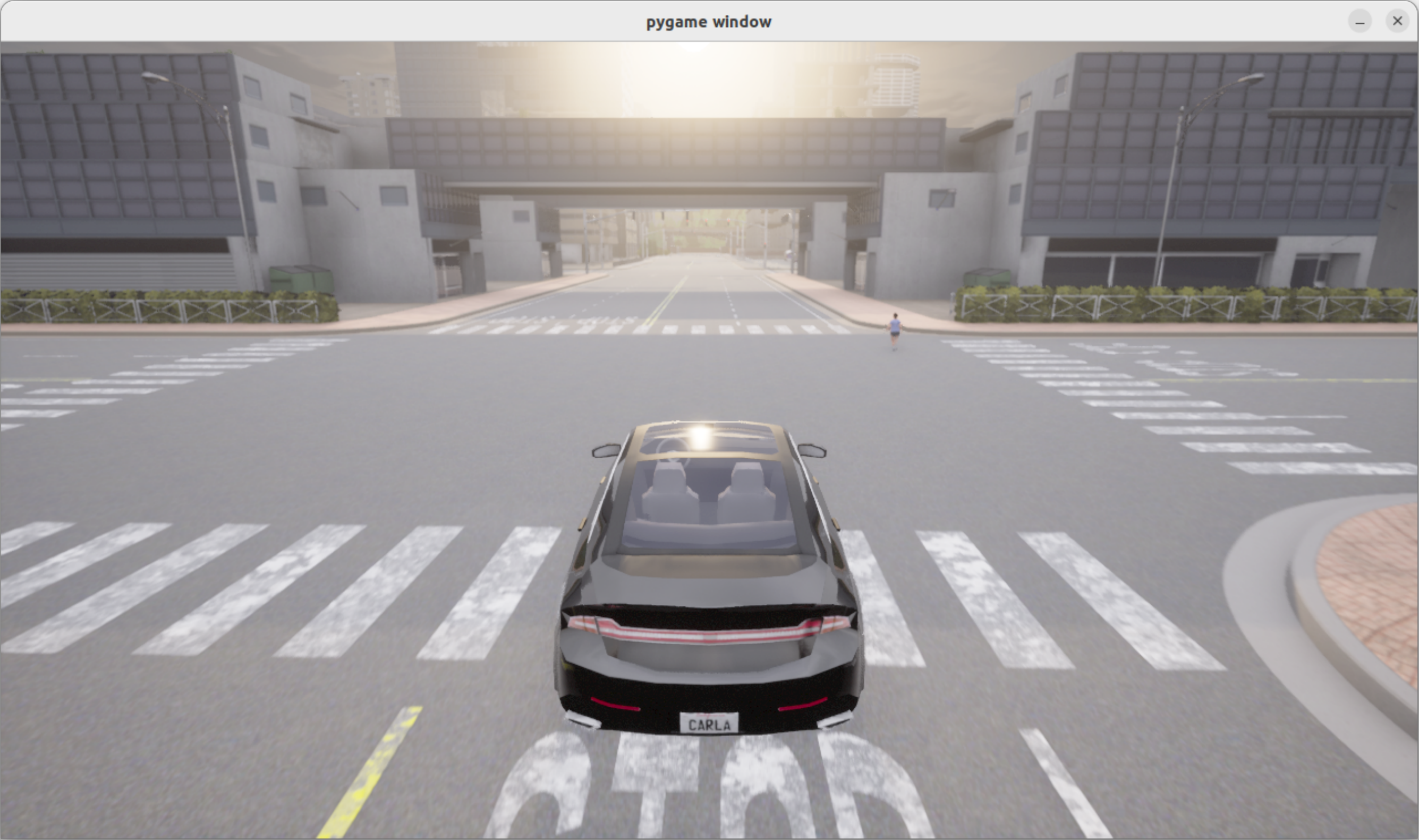}
    \caption{Failed generation example: the pedestrian is on the wrong side of the road.}
    \label{fig:results:screenshot_failed_example_0}
\end{figure}

This code is executable, however it results in an incorrect behaviour: the pedestrian crosses the intersection parallel to the Ego car and on its right side of the road, hence there is no possibility for a collision.
In order for the phrase "and must suddenly stop to avoid collision when pedestrian crosses the crosswalk" to be relevant, the pedestrian must instead either cross perpendicularly on the near side of the intersection, or parallel on the left side (Figure~\ref{fig:results:screenshot_failed_example_0}).

The user starts a second turn of dialogue with the comment: "\textit{Make sure the pedestrian crosses in front of the ego car}".
The system generates:
\newpage
\begin{small}
\begin{Verbatim}[commandchars=\\\{\}]
param map = localPath('Scenic/tests/formats/opendrive/maps/CARLA/Town05.xodr')
param carla_map = 'Town05'
model scenic.simulators.carla.model
MODEL = 'vehicle.lincoln.mkz_2017'
EGO_INIT_DIST = [20, 25]
param EGO_SPEED = VerifaiRange(7, 10)
EGO_BRAKE = 1.0
PED_MIN_SPEED = 1.0
PED_THRESHOLD = 20
param SAFETY_DIST = VerifaiRange(10, 15)
CRASH_DIST = 5
TERM_DIST = 50
behaviour EgoBehavior(trajectory):
    flag = True
    try:
        do FollowTrajectoryBehavior(target_speed=globalParameters.EGO_SPEED, trajectory=trajectory)
    interrupt when withinDistanceToAnyObjs(self, globalParameters.SAFETY_DIST) and (ped in network.drivableRegion) and flag:
        flag = False
        while withinDistanceToAnyObjs(self, globalParameters.SAFETY_DIST + 3):
            take SetBrakeAction(EGO_BRAKE)
    interrupt when withinDistanceToAnyObjs(self, CRASH_DIST):
        terminate
intersection = Uniform(*filter(lambda i: i.is4Way or i.is3Way, network.intersections))
egoManeuver = Uniform(*filter(lambda m: m.type is ManeuverType.LEFT_TURN, intersection.maneuvers))
egoInitLane = egoManeuver.startLane
egoTrajectory = [egoInitLane, egoManeuver.connectingLane, egoManeuver.endLane]
egoSpawnPt = OrientedPoint in egoInitLane.centerline
tempSpawnPt = egoInitLane.centerline[-1]
ego = Car at egoSpawnPt,
    with blueprint MODEL,
    with behaviour EgoBehavior(egoTrajectory)
\verbbf{ped = Pedestrian left of tempSpawnPt by 5,}
    with heading ego.heading,
    with regionContainedIn None,
    with behaviour CrossingBehavior(ego, PED_MIN_SPEED, PED_THRESHOLD)
require EGO_INIT_DIST[0] <= (distance to intersection) <= EGO_INIT_DIST[1]
terminate when (distance to egoSpawnPt) > TERM_DIST
\end{Verbatim}
\end{small}

\begin{figure}[h!]
    \centering
    \includegraphics[width=0.8\textwidth]{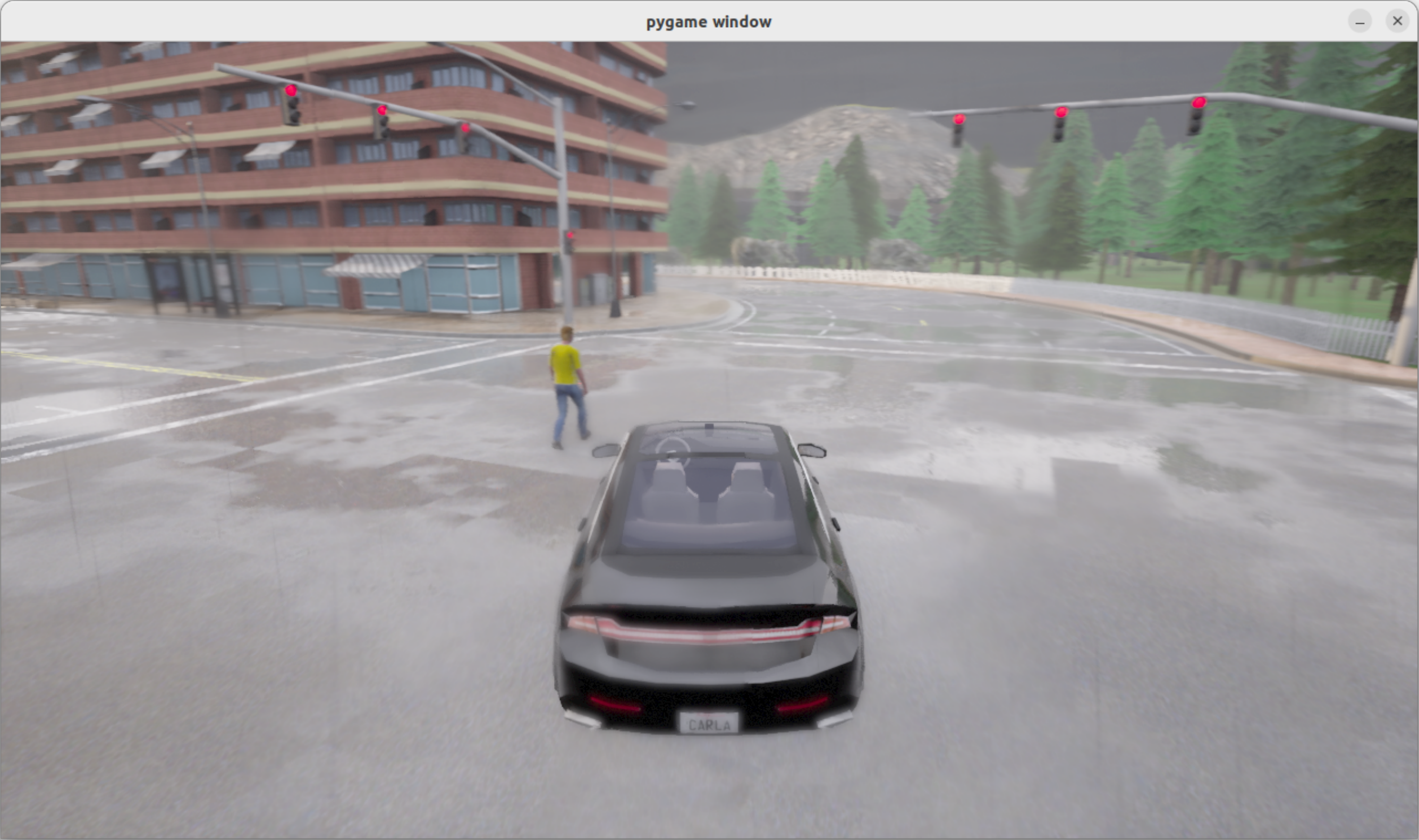}
    \caption{Failed generation example: the Ego car collides with the pedestrian, who is also incorrectly positioned.}
    \label{fig:results:screenshot_failed_example_1}
\end{figure}

The only difference here is the starting position of the pedestrian.
In this scenario the Ego car does indeed stop for some time while the pedestrian is crossing the intersection, but restarts too soon, which in most cases results in a collision.
Moreover, the pedestrian starts and ends the crossing on the road rather than on the sidewalk (Figure~\ref{fig:results:screenshot_failed_example_1}).

\newpage
The user attempts to correct the behaviour with the comment: "\textit{After braking, the ego car should wait for the pedestrian to clear the intersection before starting to move again}".
The system generates:

\begin{small}
\begin{Verbatim}[commandchars=\\\{\}]
param map = localPath('Scenic/tests/formats/opendrive/maps/CARLA/Town05.xodr')
param carla_map = 'Town05'
model scenic.simulators.carla.model
MODEL = 'vehicle.lincoln.mkz_2017'
EGO_INIT_DIST = [20, 25]
param EGO_SPEED = VerifaiRange(7, 10)
EGO_BRAKE = 1.0
PED_MIN_SPEED = 1.0
PED_THRESHOLD = 20
param SAFETY_DIST = VerifaiRange(10, 15)
CRASH_DIST = 5
TERM_DIST = 50
behaviour EgoBehavior(trajectory):
    flag = True
    try:
        do FollowTrajectoryBehavior(target_speed=globalParameters.EGO_SPEED, trajectory=trajectory)
    interrupt when withinDistanceToAnyObjs(self, globalParameters.SAFETY_DIST) and (ped in network.drivableRegion) and flag:
        flag = False
        while withinDistanceToAnyObjs(self, globalParameters.SAFETY_DIST + 3):
            take SetBrakeAction(EGO_BRAKE)
        \verbbf{while not withinDistanceToAnyObjs(self, globalParameters.SAFETY_DIST + 3):}
            \verbbf{take SetThrottleAction(0.5)}
    interrupt when withinDistanceToAnyObjs(self, CRASH_DIST):
        terminate
intersection = Uniform(*filter(lambda i: i.is4Way or i.is3Way, network.intersections))
egoManeuver = Uniform(*filter(lambda m: m.type is ManeuverType.LEFT_TURN, intersection.maneuvers))
egoInitLane = egoManeuver.startLane
egoTrajectory = [egoInitLane, egoManeuver.connectingLane, egoManeuver.endLane]
egoSpawnPt = OrientedPoint in egoInitLane.centerline
tempSpawnPt = egoInitLane.centerline[-1]
ego = Car at egoSpawnPt,
    with blueprint MODEL,
    with behaviour EgoBehavior(egoTrajectory)
ped = Pedestrian left of tempSpawnPt by 5,
    with heading ego.heading,
    with regionContainedIn None,
    with behaviour CrossingBehavior(ego, PED_MIN_SPEED, PED_THRESHOLD)
require EGO_INIT_DIST[0] <= (distance to intersection) <= EGO_INIT_DIST[1]
terminate when (distance to egoSpawnPt) > TERM_DIST
\end{Verbatim}
\end{small}

\begin{figure}[h!]
    \centering
    \includegraphics[width=0.8\textwidth]{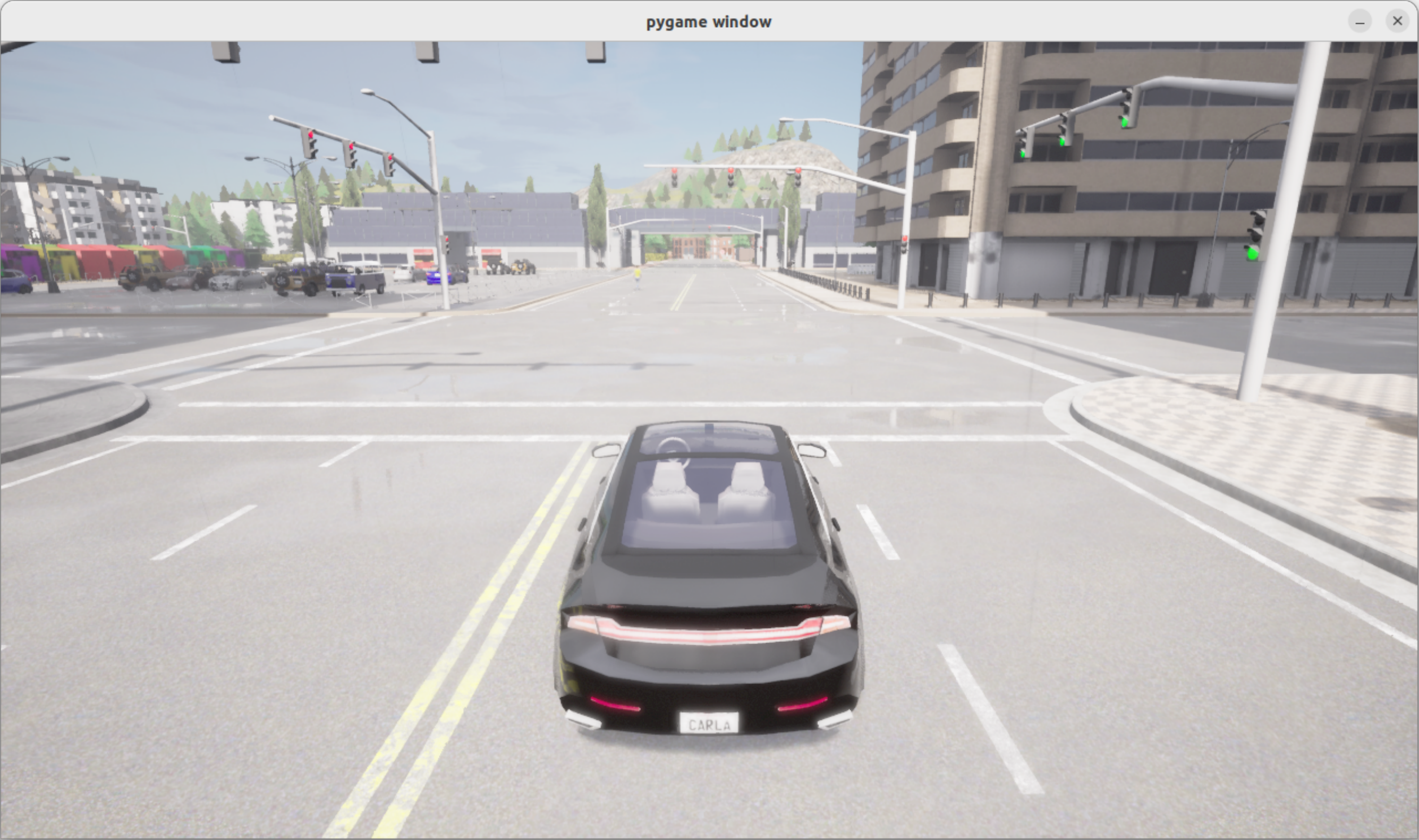}
    \caption{Failed generation example: the Ego car stops but fails to restart even though the pedestrian is far away (and still incorrectly positioned inside the road rather than on the sidewalk).}
    \label{fig:results:screenshot_failed_example_2}
\end{figure}

Now the Ego car does stop but fails to restart and keeps waiting forever (Figure~\ref{fig:results:screenshot_failed_example_2}).

The user now attempts to focus on the position of the pedestrian before attempting to fix the Ego car behaviour by specifying in two different turns of dialogue "\textit{Make sure that the pedestrian starts crossing from a sidewalk and ends crossing at a sidewalk}" and "\textit{Specify the pedestrian position as an offset from the lane}", but the system fails to fix the problem, in fact it gets "confused" and fails to even generate an executable scenario in the last turn.
Since we exhausted the stipulated number of turns, we consider this scenario as a failure.

\end{document}